\pdfoutput=1
\documentclass[11pt,dvipsnames]{article}

\usepackage[final]{acl}

\usepackage{times}
\usepackage{latexsym}

\usepackage[T1]{fontenc}
\usepackage[utf8]{inputenc}

\usepackage{microtype}

\usepackage{inconsolata}

\usepackage[inline]{enumitem}

\usepackage{graphicx}
\usepackage{tabularx}
\usepackage{multirow}

\usepackage[breakable,skins]{tcolorbox}
\usepackage{listings}

\usepackage{amsmath}

\newtcolorbox{promptbox}[1][]{
    colback=white,
    colframe=white,
    boxrule=0pt,
    left=0pt,
    right=0pt,
    top=-8pt,
    bottom=-4pt,
    enhanced
}

\newtcolorbox{todobox}[1][]{
    colback=yellow!40,
    colframe=yellow!50,
    boxrule=0.2pt,
    title={\textcolor{black}{TODO}},
    colbacktitle=yellow!10!white,
    fonttitle=\upshape\bfseries,
    arc=8pt,
    left=6pt,
    right=6pt,
    top=4pt,
    bottom=4pt,
    breakable,
    enhanced
}
\newtcolorbox{infobox}[1][]{
    colback=gray!40,
    colframe=gray!50,
    boxrule=0.2pt,
    title={\textcolor{black}{INFO}},
    colbacktitle=gray!10!white,
    fonttitle=\upshape\bfseries,
    arc=8pt,
    left=6pt,
    right=6pt,
    top=4pt,
    bottom=4pt,
    breakable,
    enhanced
}

\usepackage{array} % For custom column definitions

\usepackage[normalem]{ulem}
\newcommand{\meanci}[2]{#1 \textcolor{darkgray}{\pm #2}}
\newcommand{\meancimb}[2]{#1 \textcolor{darkgray}{\pm #2}}
\newcommand{\meancimbs}[2]{\underline{#1} \textcolor{darkgray}{\pm #2}}
\newcommand{\meancimbstwo}[2]{\dashuline{#1} \textcolor{darkgray}{\pm #2}}
\newcommand{\meanciobs}[2]{\textbf{\underline{#1}} \textcolor{darkgray}{\pm #2}}

\newcommand{\gptThree}{\texttt{GPT-3.5}}
\newcommand{\gptFourO}{\texttt{GPT-4o}}
\newcommand{\llama}{\texttt{Llama-3.3}}
\newcommand{\deepseek}{\texttt{DeepSeek-R1}}
\newcommand{\oOne}{\texttt{o1-mini}}

\newcommand{\FS}{\texttt{FS}}
\newcommand{\PF}{\texttt{PF}}
\newcommand{\RS}{\texttt{RS}}
\newcommand{\ZS}{\texttt{ZS}}

\newcommand{\BLEU}[0]{\textsc{Bleu}}
\newcommand{\lgh}{Ladin (Gherd\"eina)}
\newcommand{\lvb}{Ladin (Val Badia)}
\newcommand{\ita}{Italian}
\newcommand{\gh}{Gherd\"eina}
\newcommand{\vb}{Val Badia}
\newcommand{\itshort}{IT}
\newcommand{\ghshort}{GH}
\newcommand{\vbshort}{VB}
\newcommand{\tto}{{$\rightarrow$}}

\newcommand{\srcf}[1]{\textcolor{red}{\textbf{#1}}}
\newcommand{\midf}[1]{\textcolor{black}{\textbf{#1}}}
\newcommand{\tgtf}[1]{\textcolor{blue}{\textbf{#1}}}

\newcommand{\new}[1]{#1}

\usepackage{multirow}

\usepackage{tikz}
\usetikzlibrary{positioning, shapes, backgrounds, shapes.multipart}

\usepackage{pgfplots}
\usepgfplotslibrary{statistics}
\pgfplotsset{compat=1.8}
\definecolor{ibmblue}{RGB}{69,137,255}  
\definecolor{ibmgreen}{RGB}{66,190,101}  
\definecolor{ibmorange}{RGB}{255,131,43}  

\usepackage{booktabs}

\usepackage{soul}

\newcommand{\correct}[1]{\sethlcolor{green!30} \underline{\smash{\hl{#1}}}}
\newcommand{\frex}[1]{\sethlcolor{ibmblue!16} \underline{\smash{\hl{#1}}}}

\title{Compensating for Data with Reasoning:\\Low-Resource Machine Translation with LLMs}
\author{Samuel Frontull \and Thomas Ströhle \\
Department of Computer Science / University of Innsbruck \\
\texttt{\{samuel.frontull,thomas.stroehle\}@uibk.ac.at}}

\begin{document}
\maketitle
\begin{abstract}
Large Language Models (LLMs) have demonstrated strong capabilities in multilingual machine translation, sometimes even outperforming traditional neural systems.
However, previous research has highlighted the challenges of using LLMs --- particularly with prompt engineering --- for low-resource languages.
In this work, we introduce Fragment-Shot Prompting, a novel in-context learning method that segments input and retrieves translation examples based on syntactic coverage, along with Pivoted Fragment-Shot, an extension that enables translation without direct parallel data.
We evaluate these methods using GPT-3.5, GPT-4o, o1-mini, LLaMA-3.3, and DeepSeek-R1 for translation between Italian and two Ladin variants, revealing three key findings:
(1) Fragment-Shot Prompting is effective for translating into and between the studied low-resource languages, with syntactic coverage positively correlating with translation quality;
(2) Models with stronger reasoning abilities make more effective use of retrieved knowledge, generally produce better translations, and enable Pivoted Fragment-Shot to significantly improve translation quality between the Ladin variants;
and (3) prompt engineering offers limited, if any, improvements when translating from a low-resource to a high-resource language, where zero-shot prompting already yields satisfactory results.
We publicly release our code and the retrieval corpora on \url{https://github.com/schtailmuel/llm-lrlmt}.
\end{abstract}

\section{Introduction}

In recent years, LLMs have made significant advancements in machine translation, gaining widespread attention and achieving promising results, especially for high-resource languages~\cite{Zhang:etal:2023}. However, LLMs often face challenges when applied to low-resource scenarios where limited training data and resources are available, leading to poor translation quality~\cite{Robinson:etal:2023,Hendy:2023,Bawden:etal:2023}. In particular, LLMs have limited awareness of (different variants of) smaller languages and struggle to distinguish between them and in producing coherent output~\cite{court-elsner-2024-shortcomings,Ondrejova:etal:2024}. 
In contrast, fine-tuning specialised models can be a more effective approach. However, \citet{Gu:etal:2018} found that fewer than 13,000 sentence pairs are not enough to train a neural machine translation model to an acceptable quality. Therefore, methods that can better exploit the potential of limited data are particularly in demand, and LLMs are a promising solution due to their strong generalisation capability.
Previous studies have explored different techniques to improve LLM performance for low-resource languages~\cite{Elsner:Needle:2023, zhang-etal-2024-teaching, merx:etal:2024, guo-etal-2024-teaching, Gao:etal:2024, Shu:etal:2024, Moslem:etal:2023,Bawden:etal:2023}. These approaches typically enhance LLM output by incorporating additional information, such as dictionaries, grammatical rules, or example sentences via Retrieval Augmented Generation (RAG)~\cite{Lewis:etal:2020}.
Although LLMs still lag behind traditional neural translation systems in the translation of low-resource languages~\cite{Robinson:etal:2023}, they hold significant potential for further exploration, especially as they continue to evolve. Recent advances in the multi-hop reasoning capabilities of LLMs have opened up new possibilities, especially in low-resource scenarios.

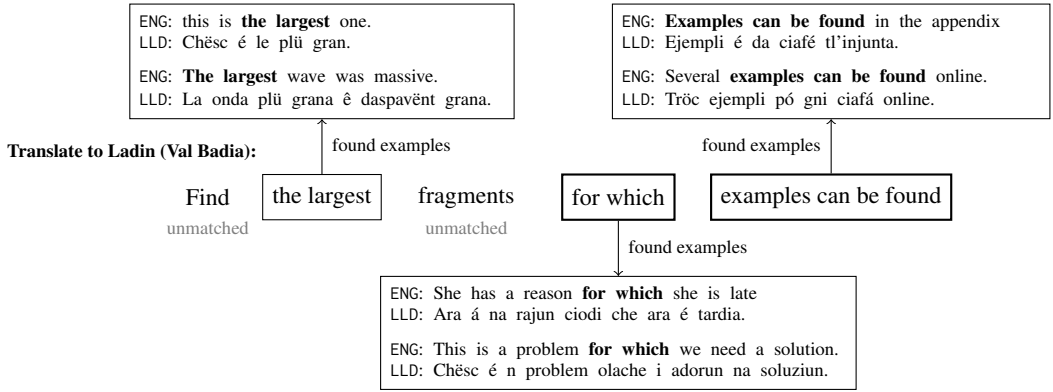
\begin{figure*}[t]
    \centering
    \begin{tikzpicture}        
        % \tikzset{every node/.style={font=\small\sffamily}}
        % Tokenized words
        \node[] (w0) at (-6, 3.6) {\scriptsize \textbf{Translate to Ladin (Val Badia):}};
        \node[rectangle, minimum height=.6cm] (w1) at (-5, 3) {\small Find};
        \node[rectangle, draw=black, minimum height=.6cm] (w2) at (-3.5, 3) {\small the largest};
        \node[rectangle, minimum height=.6cm] (w3) at (-1.6, 3) {\small fragments};
        \node[rectangle, draw=black, thick, minimum height=.6cm] (w4) at (0.4, 3) {\small for which};
        \node[rectangle, thick, draw=black, minimum height=.6cm] (w5) at (3.2, 3) {\small examples can be found};

        \node[gray] at (-1.6, 2.6) {\scriptsize unmatched};
        \node[gray] at (-5, 2.6) {\scriptsize unmatched};
        
        %\node[draw=black, dashed, rectangle, text width=3.8cm] (ex1) at (-5, 1.2) {\scriptsize
        %\textbf{find}:\\[2pt]
        % - (VERB) trovare, scoprire, individuare\\[-4pt]
        % - (NOUN) scoperta, ricerca
        %- (VERB) ciafé, abiné, araté, miné\\[-4pt]
        %- (NOUN) descurida, inrescida
        %};
        %\draw[->, black] (w1.south) -- (ex1.north) node[midway, left] {\scriptsize found (Dictionary)};
        
        \node[draw=black, rectangle, text width=4.8cm] (ex3) at (-3.5, 4.8) {\scriptsize
            \texttt{ENG}: this is \textbf{the largest} one.\\
            % \texttt{IT}: Questo è il più grande.\\[5pt]
            \texttt{LLD}: Ch\"esc é le plü gran.\\[5pt]
            \texttt{ENG}: \textbf{The largest} wave was massive.\\[-5pt]
            % \texttt{IT}: L'onda più grande era enorme.
            \texttt{LLD}: La onda plü grana ê daspavënt grana.
        };
        \draw[->, black] (w2.north) -- (ex3.south) node[midway, right] {\scriptsize found examples};

        \node[draw=black, rectangle, text width=6cm] (ex2) at (.4, 1.2) {\scriptsize
            \texttt{ENG}: She has a reason \textbf{for which} she is late\\
            % \texttt{IT}: Ha una ragione per cui è in ritardo.\\[5pt]
            \texttt{LLD}: Ara á na rajun ciodi che ara é tardia.\\[5pt]
            \texttt{ENG}: This is a problem \textbf{for which} we need a solution.\\
            % \texttt{IT}: Questo è un problema per cui abbiamo bisogno di una soluzione.\\[5pt]
            \texttt{LLD}: Ch\"esc é n problem olache i adorun na soluziun.\\
        };
        \draw[->, black] (w4.south) -- (ex2.north) node[midway, right] {\scriptsize found examples};

        \node[draw=black, rectangle, text width=5.5cm] (ex4) at (3.2, 4.8) {\scriptsize
            \texttt{ENG}: \textbf{Examples can be found} in the appendix\\
            % \texttt{IT}: Esempi possono essere trovati nell'appendice.\\[5pt]
            \texttt{LLD}: Ejempli é da ciafé tl'injunta.\\[5pt]
            \texttt{ENG}: Several \textbf{examples can be found} online.\\[-5pt]
            \texttt{LLD}: Tröc ejempli pó gni ciafá online.
            % \texttt{IT}: Diversi esempi possono essere trovati online.
        };
        \draw[->, black] (w5.north) -- (ex4.south) node[midway, left] {\scriptsize found examples};
    \end{tikzpicture}
    \caption{Fragment-Shot Prompting}
    \label{fig:fragments-prompting}
\end{figure*}

This work aims to evaluate the effectiveness of different prompting techniques for machine translation to and from the low-resource language Ladin using LLMs, in the case of Italian and the two standard variants of Ladin: Val Badia and \gh{}. Rather than fine-tuning LLMs~\cite{yong-etal-2023-bloom,zhu-etal-2024-fine,stap-etal-2024-fine,toraman-2024-adapting,vieira-etal-2024-much}, our approach aims to stimulate the generalization capabilities of LLMs through \emph{In-Context Learning} (ICL)~\cite{Rubin:etal:2022,cahyawijaya-etal-2024-llms,dong-etal-2024-survey}, using a single RAG-augmented prompt. 
Specifically, it explores what can be achieved with a small set of parallel sentences available for as retrieval corpus.

Our key contributions:
\begin{enumerate*}[label=($\roman*$)]
    \item We introduce the \emph{Fragment-Shot} prompting technique, a novel prompting method that offers exemplary translations for individual fragments of the input sentence, selected to ensure broad syntactic coverage (Figure~\ref{fig:fragments-prompting}). Furthermore, we extend this approach with the \emph{Pivoted Fragment-Shot} method, which enables translation between two languages that lack direct parallel data by leveraging a pivot language. 
    \item We evaluate the performance of \gptThree{}, \gptFourO{}, \oOne{}, \llama{}, and \deepseek{} on translation tasks between two variants of Ladin and Italian using four prompting methods: zero-shot, random-shot, Fragment-Shot, and Pivoted Fragment-Shot. We further examine the role of LLM reasoning capabilities in low-resource language translation through a coverage correlation analysis, assessing the relationship between translation quality and retrieved reference data, as well as a qualitative evaluation of the results.
    \item We publicly release our code along with the retrieval datasets containing parallel sentences for \gh{}--Italian and for \vb{}--\gh{}, to support further research on Ladin and low-resource languages in general.
\end{enumerate*}
These contributions seek to illustrate how LLMs can be leveraged in low-resource settings and deepen our understanding of their reasoning capabilities.

\section{Related Work}
% check: https://github.com/hsing-wang/Awesome-LLM-MT
The use of LLMs for machine translation has emerged as an active research area at the latest since the release of ChatGPT~\cite{Zhang:etal:2023}. Researchers have increasingly explored the potential of LLMs in comparison to traditional neural machine translation (NMT) systems, showing that human annotators, in some cases, preferred ChatGPT over mainstream NMT systems~\cite{Manakhimova:etal:2023}. 
However, the way LLMs are prompted plays a critical role and affects translation quality~\cite{Zhang:etal:2023,Agrawal:etal:2023,Vilar:etal:2023}.
Moreover, there is experimental evidence showing that GPT-models underperform for low-resource and African languages~\cite{Robinson:etal:2023}.

\paragraph{LLM-MT with RAG and Prompt Engineering}

The \emph{zero-shot} approach~\citep{Robinson:etal:2023} is the simplest way to prompt an LLM for translation, relying solely on the model’s inherent language understanding without task-specific examples. However, translating low-resource languages requires more sophisticated prompt engineering techniques: Some of the most effective strategies include Few-Shot Prompting~\citep{openai:gpt3:2020}, which improves output quality by providing a few illustrative examples; RAG~\citep{Lewis:etal:2020}, which enriches prompts with relevant external information to reduce hallucinations and enhance translation quality; and Chain-of-Thought Prompting~\citep{Wei:etal:2022}, which enables models to tackle complex problems by decomposing them into smaller, sequential reasoning steps. 
Several studies have recently focused on improving LLM performance for low-resource languages through prompt engineering and/or RAG~\cite{Elsner:Needle:2023,zhang-etal-2024-teaching,merx:etal:2024,guo-etal-2024-teaching}. Various strategies that enrich the prompt with supplementary information have been shown to improve translation quality. Examples include
\begin{enumerate*}[label=($\roman*$)]
    \item \emph{random-shot}, where randomly selected translation pairs are provided in the prompt~\cite{Zhang:etal:2023}, 
    \item \emph{dictionary-prompting}, where dictionary entries or word definitions are included~\citep{merx:etal:2024, zhang-etal-2024-teaching, Elsner:Needle:2023, guo-etal-2024-teaching}, as well as 
    \item the inclusion of translations of similar sentences in the prompt~\citep{merx:etal:2024}.
\end{enumerate*}
The Fragment-Shot and Pivoted Fragment-Shot we present in this work build on these ideas. In contrast to \citet{merx:etal:2024}, we base our method on the highest word overlap, as semantic similarity via embeddings is not feasible due to the unavailability of suitable models.

\paragraph{Machine Translation and Multi-Hop Reasoning}
There is strong evidence of latent multi-hop reasoning in LLMs~\citep{Yang:etal:2024}, a capability that enables models to draw on multiple pieces of information --- potentially from different parts of a prompt or even from external knowledge --- to arrive at a final answer.
\citet{Puduppully:etal:2023} applied this idea in DecoMT, a decomposed prompting method that significantly outperformed standard few-shot approaches, particularly for translation between related low-resource languages.
This method shares similarities with our Fragment-Shot approach. However, unlike the two-stage process that first segments the text and then translates each part with added context, our method translates the full text in a single prompt.
Furthermore, it raises important questions about the performance of newer reasoning models on low-resource languages and the effective evaluation of the reasoning process involved in translation.

\paragraph{Machine Translation for Ladin}
To date, only two studies have explicitly focused on Ladin in the context of machine translation, both relying on basic zero-shot prompting without exploring more advanced prompting strategies: \citet{Frontull:Moser:2024} explored the effect of different models used for back-translation, including \gptThree{}. Similarly, \citet{Valer:etal:2024} introduced a bidirectional machine translation system for Fassa Ladin, highlighting the benefits of multilingual training and knowledge transfer from related languages like Friulian and compared the results to the ones produced by \gptFourO{}.

\begin{figure*}[t]
\lstset{
    basicstyle=\footnotesize\ttfamily,
    keywordstyle=\bfseries,
    showstringspaces=false,
    escapeinside={(*@}{@*)},
    numbers=left,
    xleftmargin=2em,
    framexleftmargin=2em,
    numberstyle=\tiny\color{gray},
    frame=tb
}

\begin{tikzpicture}[overlay]
  \node[anchor=east,text width=11em, align=left,draw=none, dashed] (think) at (14.7,-1.6) {
  \scriptsize\texttt{\textbf{\deepseek{}}\\[4pt]
      \begin{tabular}{@{}l@{}}
      ...**\srcf{suvënz}**: The example "l tlama suvënz \\
      te ustaria" translates to "frequenta spesso\\
      il bar". In Val Badia, "\midf{spesso}" is "\tgtf{gonot}" \\
      as seen in "la uma bruntora gonot le möt" \\
      (la mamma rimprovera spesso il bambino). \\
      So "\srcf{suvënz}" (often) becomes "\tgtf{gonot}"...
    \end{tabular}
    }
  };
  \draw (think.south west) -- (think.south east);

\node[anchor=east,text width=11em, align=left] (think2) at (14.7,-3.6) {
  \scriptsize\texttt{\textbf{\llama{}}\\[4pt]
      \begin{tabular}{@{}l@{}}
      ...**\srcf{suvënz}** means often, \\
      so "{spesso}" or a similar term \\
      in Ladin (Val Badia) would work...
    \end{tabular}
    }};
  \draw (think2.south west) -- (think2.south east);

\node[anchor=east,text width=11em, align=left] (think3) at (14.7,-5.4) {
  \scriptsize\texttt{\textbf{\gptFourO{}}\\[4pt]
      \begin{tabular}{@{}l@{}}
      ...The word "\srcf{suvënz}" is seen in Gherdëina\\ 
      as "\midf{spesso}" in Italian, with no alternative \\
      translation provided in Val Badia,\\
      suggesting that "sovenz" might be retained...
    \end{tabular}
    }};     
\end{tikzpicture}

\begin{lstlisting}
Translate the following sentence from 
Ladin (Gherdëina) into Ladin (Val Badia):

>>I pësc mor (*@\srcf{suvënz}@*) per gauja dla cunzentrazion
    auta de tuesse tl'eghes.<<
...

  Examples that illustrate the usage of **(*@\srcf{suvënz}@*)**:

  - Ladin (Gherdëina): l tlama (*@\srcf{suvënz}@*) suvënz te ustaria
  - Italian: frequenta (*@\midf{spesso}@*) il bar

    Examples that illustrate the usage of **(*@\midf{spesso} il@*)**:

    - Italian: la mamma rimprovera (*@\midf{spesso}@*) il bambino
    - Ladin (Val Badia): la uma bruntora (*@\tgtf{gonot}@*) le möt
...
\end{lstlisting}
\caption{Example of Pivoted-Fragments Prompting and the corresponding reasoning employed by different LLMs.}
% \vspace{-0.3cm}
\label{fig:pivoted-prompting}
\end{figure*}

\section{Prompting Techniques}

This section details the four prompting methodologies applied in our experiments to enhance machine translation performance for Ladin. 

\paragraph{Zero-Shot (\ZS)}

The \emph{zero-shot} method~\cite{Robinson:etal:2023, Gao:etal:2024,Hendy:2023,Bawden:etal:2023} relies solely on the model's pre-existing knowledge. The prompt directly instructs the model to translate sentence into the target, without providing explicit translation examples or lexical guidance. This baseline approach tests the model's intrinsic understanding of Ladin syntax and vocabulary.

\paragraph{Random-Shot (\RS)}

In the \emph{random few-shot} technique~\cite{Agrawal:etal:2023,Robinson:etal:2023,Bawden:etal:2023} we provided $16$ randomly selected source--target language translation pairs followed by the sentence to translate. 
These examples, even if not necessarily related, serve as in-context references, encouraging the model to infer translation and language patterns.

\paragraph{Fragment-Shot (\FS)}

In the \emph{Fragment-Shot} method, the sentence to be translated is partitioned into contiguous word sequences which we call \emph{fragments}. 
For each fragment, we retrieve validated translation examples from the training corpus in which the fragment appears in the source sentence (see Figure~\ref{fig:fragments-prompting}). 
These fragments are constructed based on their occurrences in the retrieval data. We start with a sliding window of seven contiguous words and check whether any sentence in the retrieval data contains an exact match for this fragment on the source side. If such examples are found, we randomly select up to six sentence pairs containing the full fragment and include them in the prompt.
If no matches are found, the window size is progressively reduced until we arrive at one-word units. 
Also, the fragments are chosen to avoid overlap, ideally ensuring a complete but non-redundant coverage of the sentence. We prioritize examples that explicitly contain the fragment in the source sentence, rather than selecting on the basis of a global sentence similarity. The examples serve to illustrate plausible translations of the fragment in context.
To allow reproducibility and further experimentation, we released a Python package\footnote{\url{https://pypi.org/project/fragmentshot/}} that implements this method.

\paragraph{Pivoted Fragment-Shot (\PF)} The \emph {pivoted} Fragment-Shot approach extends the \FS-approach by enabling translation between two languages for which no direct parallel data is unavailable, leveraging a pivot language. 
This method applies the \FS{} method in a nested manner across two different bilingual corpora: source-pivot and pivot-target. 
In our case we implemented this method for translating between \lgh{} and \lvb{} via \ita{}. 
Specifically, for a given sentence in the source language, we extract fragments as in the \FS{}-method. For each fragment, we search the source-pivot corpus for up to three sentence pairs that contain the fragment on the source side and the corresponding pivot translations. We reduced to $2$ for that exceeded the context size of the model. We then treat the pivot translations as new source texts and perform a second round of this search in the pivot-target corpus for (again, up to three) sentence pairs in the pivot language that contain fragments of the pivot translation, along with their translations into the target language. 
Given the substantial overlap between the Italian sentences in both corpora, we deliberately excluded exact pivot-sentence matches in order to force reasoning about fragments and thus simulate a more difficult translation problem.
%
%It shows a \lgh{} sentence that contains the fragment \srcf{suvënz} for which an example can be found in the parallel corpus (sentence at line 10). The corresponding italian translation contains the fragment \emph{spesso il} for which, again, an example can be found in the \ita{}--\vb{} parallel corpus with the corresponding translation in \lvb{} (line 16). From these examples, the LLMs should infer which part of this sentence correspond to the translation initial fragment in \gh{}, which in this case would be \tgtf{gonot}. The right side illustrates the reasoning processes employed by \deepseek{}, \llama{}, and \gptFourO{} when translating the text shown in the prompt to the left for the fragment \emph{suvënz} in \gh{}.  
%Ideally, this nesting should make it possible to derive the corresponding translation of the original fragments in the target language.
%
Figure~\ref{fig:pivoted-prompting} illustrates this approach by showing, on the left, the examples provided for the fragment \srcf{suvënz} occurring in the \gh{} sentence, which connects via Italian to the corresponding \vb{} translation \tgtf{gonot}. The right side illustrates the reasonings employed by \deepseek{}, \llama{}, and \gptFourO{}.  

\section{The Ladin Language}

Ladin is a Rhaeto-Romance language spoken in the Dolomite region of Northern Italy. Ladin is characterised by its internal linguistic diversity, with five main regional variants: \emph{Val Badia}, \emph{Gherdëina}, \emph{Fassa}, \emph{Fodom}, and \emph{Anpezo}. Each variant has its own orthographic conventions, vocabulary, and grammatical structures, making it a compelling case for machine translation research. 
%The main challenge in developing machine translation systems for Ladin lies in creating solutions that work across these variants. 
%
This study analyses the performance of various LLMs in translating texts between Italian and the written standards of Val Badia and Gherdëina. These standards represent the Ladin varieties spoken by around 20,000\footnote{This number is based on data from the 2024 South Tyrolean Language Group Census, 2023 published by ASTAT at \url{https://astat.provinz.bz.it/de/publikationen/ergebnisse-sprachgruppenzahlung-2024}.} people in these two South Tyrolean valleys.
Due to the very limited amount of machine-readable data available for Ladin and the fact that ita variants are not distinguishable in ISO 639-3\footnote{\url{https://iso639-3.sil.org}} it is likely that LLMs have minimal exposure to Ladin and are unaware of its internal variation.
To give an intution on the similarity between the two variants and Italian and on the difficulty of the translation task, we computed the \BLEU{} score obtained by leaving the text untranslated, which resulted in a score of $12.9$ for \vb{}--\gh{}, $5.0$ for \ita{}--\vb{} and $4.3$ for \ita{}--\gh{}.
\paragraph{Retrieval Corpora} As retrieval corpora, we have included the following datasets: $18,140$ sentences for \vb{}--\ita{}, which have already been published\footnote{\url{https://doi.org/10.57967/hf/1878}}, and $19,971$ sentences for \gh{}--\ita{}, extracted from the dictionary \lgh{}--Italiano~\cite{Forn:2013}. \new{Since the Italian sentences of both datasets largely overlap, we aligned them to construct an additional parallel dataset for \vb{}--\gh{} with $14,953$ sentences}. We make both the \gh{}--\ita{}\footnote{\url{https://huggingface.co/datasets/sfrontull/lld_gherd-ita}} and \vb{}–\gh{}\footnote{\url{https://huggingface.co/datasets/sfrontull/lld_valbadia-lld_gherd}} datasets publicly available under a \texttt{CC BY-NC-SA 4.0} license. %\footnote{\url{https://creativecommons.org/licenses/by-nc-sa/4.0}} license.
These sentences, originally created as language reference material, are relatively short and simple with an average length of $\approx 25$ characters. 
\paragraph{Test Data} For this study, we had $175$ sentences from the FLORES+~\cite{nllb-24} dataset (dev split) translated into Val Badia and Gherdëina. These translations were produced by native Ladin speakers and professional translators affiliated with the Ladin Cultural Institute ``Micurá de Rü''.%\footnote{\url{https://www.micura.it}}. 
All translators followed the guidelines provided by OLDI\footnote{\url{https://oldi.org/guidelines}},
%the Open Language Data Initiative (OLDI)\footnote{\url{https://oldi.org/guidelines}}, 
ensuring consistency and accuracy in the translation process. 
%The reference translations were based on English texts.

\begin{table*}[t]
\vspace{-0.3cm}
    \centering 
    \small
    % \hline
    % \textbf{English $\rightarrow$ Italian}  & & & & & \\
    % zero-shot &	$30.25/59.68$ &	$31.31/60.33$ & &	& $30.35/59.60$\\
    \begin{tabularx}{\textwidth}{p{3.8cm} p{.6cm}rrrrr}
    \toprule
Translation direction / \BLEU{} & & \textbf{\gptThree{}} & \textbf{\gptFourO{}} & \textbf{\oOne{}}& \textbf{\llama{}} & \textbf{\deepseek{}} \\
%& & \scriptsize{BLEU } & \scriptsize{BLEU} & \scriptsize{BLEU} & \scriptsize{BLEU} & \scriptsize{BLEU}\\
% \midrule
% \textbf{English \tto{} Italian} & \texttt{\ZS} &	$(27.21,28.87)$ &	 &	 &	$(27.09,28.73)$ &  \\
\midrule
\textbf{Val Badia \tto{} Italian} 
  & \texttt{\ZS} 
  & $\meanci{19.35}{2.09}$ 
  & $\meancimb{22.90}{2.11}$ 
  & $\meanci{18.36}{2.31}$ 
  & $\meanci{20.31}{2.10}$ 
  & $\meanci{22.47}{2.04}$ \\
  
  & \texttt{\RS} 
  & $\meancimb{20.75}{2.07}$ 
  & $\meanci{22.83}{2.11}$ 
  & $\meanci{18.29}{2.09}$ 
  & $\meanci{20.44}{2.27}$ 
  & $\meancimb{22.80}{2.05}$ \\
   
  & \texttt{\FS} 
  & $\meanci{19.77}{2.01}$ 
  & $\meanci{22.49}{1.99}$ 
  & $\meancimbs{21.07}{2.16}$ 
  & $\meancimbs{21.98}{2.17}$ 
  & $\meanci{22.46}{1.94}$ \\
\midrule
\textbf{Gherdëina \tto{} Italian} 
  & \texttt{\ZS} & $\meanci{18.52}{2.04}$ & $\meancimb{23.03}{2.09}$ & $\meanci{17.26}{1.99}$ & $\meancimb{21.02}{2.12}$ & $\meancimb{23.02}{1.96}$ \\
  & \texttt{\RS} & $\meanci{19.35}{1.98}$ & $\meanci{22.15}{2.32}$ & $\meanci{18.63}{2.04}$ & $\meanci{20.59}{2.11}$ & $\meanci{22.45}{2.00}$ \\
  & \texttt{\FS} & $\meancimb{19.43}{1.82}$ & $\meanci{21.18}{1.95}$ & $\meancimb{19.78}{1.86}$ & $\meanci{20.96}{1.90}$ & $\meanci{21.79}{1.75}$ \\
\midrule
\textbf{Italian \tto{} Val Badia} 
  & \texttt{\ZS} 
  & $\meanci{4.91}{1.23}$ 
  & $\meanci{5.70}{1.40}$ 
  & $\meanci{4.67}{1.22}$ 
  & $\meanci{6.46}{1.29}$ 
  & $\meanci{6.31}{1.24}$ \\
  & \texttt{\RS} 
  & $\meanci{5.01}{1.18}$ 
  & $\meanci{5.70}{1.45}$ 
  & $\meanci{5.22}{1.30}$ 
  & $\meanci{7.94}{1.47}$ 
  & $\meanci{6.91}{1.38}$ \\
  & \texttt{\FS} 
  & $\meancimbs{7.28}{1.31}$ 
  & $\meancimbs{13.31}{1.63}$ 
  & $\meancimbs{11.37}{1.59}$ 
  & $\meancimbs{12.85}{1.55}$ 
  & $\meanciobs{14.22}{1.61}$ \\
  \midrule
\textbf{Italian \tto{} Gherdëina} 
  & \texttt{\ZS} 
  & $\meanci{6.73}{1.18}$ 
  & $\meanci{5.53}{1.20}$ 
  & $\meanci{6.69}{1.15}$ 
  & $\meanci{9.50}{1.35}$ 
  & $\meanci{8.77}{1.30}$ \\
  & \texttt{\RS} 
  & $\meanci{6.65}{1.15}$ 
  & $\meanci{7.56}{1.26}$ 
  & $\meanci{6.39}{1.13}$ 
  & $\meanci{9.50}{1.28}$ 
  & $\meanci{9.07}{1.23}$ \\
  & \texttt{\FS} 
  & $\meancimbs{8.58}{1.27}$ 
  & $\meancimbs{12.58}{1.50}$ 
  & $\meancimbs{11.51}{1.46}$ 
  & $\meancimbs{13.32}{1.58}$ 
  & $\meanciobs{14.63}{1.48}$ \\
  \midrule
\textbf{Val Badia \tto{} Gherdëina} 
  & \texttt{\ZS} 
  & $\meancimb{10.52}{1.41}$ 
  & $\meanci{11.15}{1.61}$ 
  & $\meanci{8.94}{1.25}$ 
  & $\meanci{15.01}{1.69}$ 
  & $\meanci{13.11}{1.52}$ \\
  & \RS 
  & $\meanci{10.39}{1.34}$ 
  & $\meanci{12.21}{1.49}$ 
  & $\meanci{12.07}{1.64}$ 
  & $\meanci{16.61}{1.83}$ 
  & $\meanci{15.28}{1.64}$\\
  & \FS 
  & $\meancimbs{13.91}{1.68}$ 
  & $\meancimbs{25.46}{2.13}$ 
  & $\meancimbs{23.81}{1.75}$ 
  & $\meancimbs{24.16}{1.99}$ 
  & $\meanciobs{28.60}{2.23}$ \\
  & \texttt{\PF} 
  & $\meanci{10.78}{1.48}$ 
  & $\meancimbstwo{16.19}{1.62}$ 
  & $\meancimbstwo{14.52}{1.70}$ 
  & $\meanci{15.52}{1.65}$ 
  & $\meancimbstwo{20.94}{1.89}$ \\
  \midrule
\textbf{Gherdëina \tto{} Val Badia} 
  & \texttt{\ZS} 
  & $\meanci{10.73}{1.53}$ 
  & $\meanci{11.17}{1.37}$ 
  & $\meanci{8.10}{1.20}$ 
  & $\meanci{12.46}{1.53}$ 
  & $\meanci{10.19}{1.34}$ \\
  & \RS 
  & $\meanci{11.25}{1.50}$ 
  & $\meanci{12.36}{1.45}$ 
  & $\meanci{10.83}{1.42}$
  & $\meanci{13.14}{1.53}$ 
  & $\meanci{13.75}{1.60}$\\
  & \FS 
  & $\meancimbs{13.99}{1.70}$ 
  & $\meancimbs{23.10}{2.02}$ 
  & $\meancimbs{22.21}{2.00}$ 
  & $\meancimbs{21.70}{2.01}$ 
  & $\meanciobs{26.27}{2.16}$\\
  & \PF 
  & $\meanci{10.08}{1.39}$ 
  & $\meancimbstwo{15.67}{1.69}$ 
  & $\meancimbstwo{13.32}{1.54}$  
  & $\meanci{14.07}{1.67}$ 
  & $\meancimbstwo{19.33}{1.70}$ \\
\bottomrule
    \end{tabularx}
  \caption{\BLEU{} mean scores and confidence intervals of GPT-3.5, GPT-4o, o1-mini, Llama-3.3, and DeepSeek-R1 across various Ladin and Italian translation pairs using different prompting methods as reported by \texttt{sacrebleu}.}
  \label{tab:results}
  \vspace{-0.3cm}
\end{table*}

\section{Large Language Models}

To evaluate the efficacy of the prompting techniques, we selected the following five state-of-the-art LLMs:
\begin{enumerate*}[label=($\roman*$)]
    \item \gptThree{} is a general-purpose language model from OpenAI's GPT-3 series with 175B parameters, released in 2022~\cite{openai:gpt3:2020}.
    \item \gptFourO{} a model by OpenAI, introduced in 2023, with 200B parameters and enhanced reasoning capabilities designed for complex problem-solving~\cite{hurst2024gpt}.
    \item \oOne{} a model by OpenAI optimised for reasoning with 50B parameters, launched in September 2024~\cite{jaech2024openai}.
    \item \llama{} is a text-only model by Meta AI, released in December 2024, featuring 70B parameters~\cite{touvron2023llamaopenefficientfoundation}.
    \item \deepseek{} is a reasoning-focused model by DeepSeek AI, introduced in January 2025, with 658B parameters~\cite{deepseek:r1:2024}.
\end{enumerate*}
The models were prompted using the API services: OpenAI API\footnote{\url{https://platform.openai.com}} for \gptThree{}, \gptFourO{}, and \oOne{}, DeepSeek API\footnote{\url{https://www.deepseek.ai/api}} for \deepseek{}, and Together Inference API\footnote{\url{https://www.together.ai}} for \llama{}. The hyperparameters were configured according to the default settings provided by each service.

\begin{table}[t]
  \centering
  \small
  \begin{tabularx}{\columnwidth}{Xrrrr}
\toprule
\emph{avg. duration} [s] & \textbf{\ZS} & \textbf{\RS} & \textbf{\FS} & \textbf{\PF} \\
\midrule
\textbf{\gptThree{}} &	$1.01$ &	$0.99$ &	$1.18$ &	$1.24$ \\
\textbf{\gptFourO{}} &	$1.63$ &	$1.78$ &	$6.85$ &	$7.34$ \\
\textbf{\llama{}} &	$1.74$ &	$2.04$ &	$6.49$ &	$9.54$ \\
\textbf{\oOne{}} &	$7.10$ &	$9.54$ &	$15.27$ &	$27.56$ \\
\textbf{\deepseek{}} &	$19.42$ &	$20.87$ &	$34.22$ &	$30.97$ \\\midrule
\emph{creation time} [s]  & $0.00$ &	$0.02$ &	$1.08$ &	$1.90$ \\
% \emph{avg. \# lines}   & $5$ &	$79$ &	$285$ &	$755$ \\
\emph{avg. \# chars}   & $247$ &	$2232$ &	$8974$ &	$24852$ \\
% TODO: zahlen nochmal prüfen (SF)
\bottomrule
  \end{tabularx}
  \caption{\label{tab:inference-times}  
  \new{Prompt statistics and average inference times}.
  }
  \vspace{-0.5cm}
\end{table}

\section{Results}

Table~\ref{tab:results} shows, for the selected LLMs \gptThree{}, \gptFourO{}, \oOne{}, \llama{} and \deepseek{}, the mean \BLEU{}~\cite{Post:2018} scores computed with \texttt{sacrebleu}\footnote{\url{https://github.com/mjpost/sacrebleu}} as well as the standard deviation observed for \ZS, \RS, \FS{} and \PF{} between \vb{}, \gh{} and \ita{}. 
We perform pairwise statistical significance tests using \texttt{sacrebleu} to assess whether differences in \BLEU{} scores between models and prompting strategies are meaningful or attributable to chance. 
We examined three aspects: (1. \underline{underlined}) we used the \FS{} approach as the baseline in the significance test to determine whether it yields the best results for each model and translation direction; (2. \dashuline{dashed underlined}) we used the \PF{} approach as a baseline to assess whether it outperforms \ZS{} and \RS{} methods for each model and for translations between low-resource languages; (3. \textbf{bold}) for each prompting approach and translation direction, we used the \deepseek{} model as the baseline to assess whether it outperforms the others. We underlined the \FS{} approach if it was significantly better than all others, and we highlighted \deepseek{} in bold if it outperformed all other models.
Table~\ref{tab:results} highlights three key findings: (1) In translations from Ladin to Italian, the \FS{} approach yields significant improvements only for \oOne{} and \llama{}. For all other models, and for \gh{} to \ita{} translations in particular, more sophisticated prompting techniques show little to no benefit; in some cases, \ZS{} prompting even delivers the best results. (2) In translations from Italian to Ladin, as well as between \vb{} and \gh{}, our \FS{} approach consistently achieves the highest performance. Additionally, the Pivoted-Fragments prompting method yielded significant translation improvements compared to \ZS{} and \RS{}, but only for reasoning models. (3) \deepseek{} consistently outperforms all other models, especially in translations from high-resource to low-resource languages and between low-resource language pairs.
\begin{table*}[t]
\vspace{-0.3cm}
  \centering
  \small
  \begin{tabularx}{\textwidth}{Xrrrrr|ccccc}
\toprule
& & \multicolumn{4}{c|}{fragment size} & \multicolumn{5}{c}{pearson correlation} \\
& \textbf{Coverage} & \textbf{1} & \textbf{2} & \textbf{3} & \textbf{4} 
& \textbf{\gptThree{}} & \textbf{\gptFourO{}} & \textbf{\oOne{}}& \textbf{\llama{}} & \textbf{\deepseek{}}\\\midrule
\texttt{\vbshort \tto{}\itshort} & $\meanci{0.88}{0.08}$ & $1559$ & $646$ & $122$ & $12$   
  & $0.04\;\,$ 
  & $0.01\;\,$ 
  & $0.02\;\,$ 
  & $0.04\;\,$ 
  & $0.07\;\,$ \\
\texttt{\ghshort \tto{}\itshort} & $\meanci{0.89}{0.06}$ & $1538$ & $730$ & $165$ & $22$ 
  & $0.10\;\,$ 
  & $0.00\;\,$ 
  & $0.06\;\,$ 
  & $0.01\;\,$ 
  & $0.03\;\,$ \\
\texttt{\itshort \tto{}\vbshort} & $\meanci{0.70}{0.12}$ & $1629$ & $491$ & $50$  & $3$  
  & $0.28^*$ 
  & $0.24^*$ 
  & $0.29^*$ 
  & $0.29^*$ 
  & $0.35^*$ \\
\texttt{\itshort \tto{}\ghshort} & $\meanci{0.71}{0.11}$ & $1628$ & $480$ & $54$  & $1$  
  & $0.42^*$ 
  & $0.34^*$ 
  & $0.41^*$ 
  & $0.35^*$ 
  & $0.42^*$ \\
\texttt{\vbshort \tto{}\ghshort} & $\meanci{0.85}{0.13}$ 
  & $1607$  
  & $621$  
  & $97$   
  & $13$  
  & $0.46^*$ 
  & $0.48^*$  
  & $0.46^*$  
  & $0.45^*$   
  & $0.53^*$ \\
\texttt{\ghshort \tto{}\vbshort} & $\meanci{0.83}{0.08}$  
  & $1602$  
  & $706$  
  & $146$   
  & $21$  
  & $0.40^*$ 
  & $0.47^*$  
  & $0.48^*$  
  & $0.43^*$  
  & $0.50^*$  \\
\bottomrule
  \end{tabularx}
  \caption{
  \new{\FS-coverage and pearson correlation \FS-coverage--\BLEU{} statistics}.
  }
  \label{tab:fs-coverage} 
  \vspace{-0.3cm}
\end{table*}
Table~\ref{tab:inference-times} presents statistics for the different methods. We compare the prompt creation times, the average prompt length (in characters), and the inference times of the various models. Our findings indicate that reasoning-oriented models generally require longer inference times. For all models, inference time tends to increase with prompt size, though not always in direct proportion. Notably, \PF-prompts are significantly larger
%-- approximately ten times the size of \RS-prompts -- 
and can quickly approach the model's context limits when processing longer sentences.
Table~\ref{tab:fs-coverage} presents the syntactic coverage analysis for the different language combinations for the \FS-method. We evaluated the input sentence coverage by counting
%, for each sentence, 
how many words could be exemplified and detracted those we assumed to be non-translatable (e.g., proper names). Moreover, we list the total number of fragments of size 1,2,3 and 4 found in the retrieval corpus. 
We observe a significant correlation between coverage and \BLEU{} score in translations from Italian to Ladin, as well as between two low-resource languages--indicating that higher coverage of relevant information leads to better translation quality. However, this correlation does not hold in translations from Ladin to Italian.
To give an insight into the effect of different prompting methods, we have included an example sentence in \gh{} in Table~\ref{tab:examples-lvb-lgh}, along with the outputs by selected models and the reference translation in \vb{}. 
We have highlighted the input fragments for which we could retrieve examples, as well as the correctly translated segments in the generated translations. 

\section{Discussion}

In the following, we discuss our main findings based on the results presented above.

\paragraph{Syntactic Coverage Correlates with Translation Quality into and between Low-Resource Languages}

In contrast to the Ladin to Italian direction, we observe substantial performance improvements when moving from \ZS{} or \RS{} to \FS{} or \PF{}.
While previous work has selected examples based on sentence-level similarity---using metrics like \BLEU{}~\cite{Agrawal:etal:2023} or semantic embeddings~\citep{merx:etal:2024,Shu:etal:2024}, our \FS{} and \PF{} approaches prioritise examples that maximise syntactic coverage, i.e., the inclusion of source words present in the input sentence. 
This strategy is particularly valuable in settings where building reliable embedding models is challenging due to limited data availability. The importance of lexical overlap was previously emphasized by~\citet{Agrawal:etal:2023}; we measure the coverage more directly by counting the number of complete source words for which example translations are available and reinforce this finding with the correlation results observed between this fragment coverage and \BLEU{} scores in the \FS{} setting. 
There is a clear trend in all models that the \FS{}-method help the models to produce more accurate translations and promotes adherence to standard spelling conventions. However, the degree of improvement varies between models. For example, compared to \ZS{}, the gain is approximately $+3.3$ \BLEU{} points for \gptThree{} and $+15.5$ for \deepseek{}.
\deepseek{} not only delivers the best performance gains, but also shows the highest correlation between syntactic coverage and translation quality, highlighting the importance of reasoning for effectively processing the prompts.

\paragraph{Reasoning Can Compensate for Data}

In the absence of direct parallel data, the Pivoted \FS{} method---which translates between \vb{} and \gh{}---offers a promising approach for low-resource language translation by leveraging nested \FS{} prompting with Italian as the pivot language.
%
% arbeit zu multi-hop reasoning zitieren und sagen dass wir damit beabsichtigen diese fähigkeit auszunützen
% 
%
Although the results achieved with this method are clearly inferior to those achieved with \FS{} on direct parallel data, the leap in comparison to \RS{} is significant for \gptFourO{}, \oOne{} and \deepseek{}.
\deepseek{} demonstrates notable strengths in processing the structured \PF{} prompts, significantly outperforming the other models.
These findings suggest that robust reasoning capabilities are crucial for effectively applying the \PF{} approach. \PF{} requires models to perform \emph{multi-hop} reasoning~\cite{Yang:etal:2024} across retrieved examples and pivot language rather than rely solely on language modeling or memorized translation patterns.
We thus conclude that such capabilities can, at least to some extent, compensate for the lack of extensive training corpora.
However, this is related to longer inference times (see Table~\ref{tab:inference-times}).
Further qualitative analysis revealed that \deepseek{} is the model with the highest proportion of syntactically valid words in the generated translations for the specific target variant, also leaving the smallest proportion of words untranslated.
Nevertheless, there are still challenges in fully capturing the vocabulary and morphology of the target variant. On average---compared to the reference translations---$7$--$9\%$ more words in the generated translations are syntactically not valid in the target language, indicating substantial room for improvement.
The observed weaknesses may be explained by several factors:
\begin{enumerate*}[label=($\roman*$)]
    \item the models do not appear to have prior knowledge of Ladin or a built-in distinction between its variants, as demonstrated by their performance in \ZS;
    \item the retrieval corpus used in prompt construction is relatively small; based on the statistics shown in Table~\ref{tab:fs-coverage}, we could expect coverage of only around $55\%$ for Gherdëina--Val Badia prompts;
    \item the reasoning is highly sensitive to the prompt. Figure~\ref{fig:pivoted-prompting} illustrates this issue: despite the prompt containing an example for the word \emph{suvënz} with its Italian and Val Badia translations, some models fail to correctly infer the translation. In this case, the prompt explicitly refers to the fragment \emph{spesso il}, which may have caused \gptFourO{} to claim that no example for \emph{spesso} exists. \llama{} understands the meaning of \emph{suvënz} but fails to generate the correct translation, while \deepseek{} succeeds, highlighting its comparatively stronger ability to perform \emph{multi-hop} reasoning~\cite{Yang:etal:2024}.
    %under the given prompting conditions.
\end{enumerate*}
Addressing these weaknesses could improve the overall effectiveness of the approach. Further improvements can be expected if language-specific resources are incorporated during the pre-training of the model, as discussed in~\citet{ye2025limoreasoning}.

%   MODEL		TGT_KS	TGT_UK	TGT
% input    :	0.3613	0.1667	-
% reference:	0.0050	0.1592	0.8357
% gpt-3.5  :	0.2538	0.2528	0.4933
% gpt-4o   :	0.1244	0.2544	0.6212
% o1-mini  :	0.2102	0.2301	0.5597
% Llama-3.3:	0.1389	0.2283	0.6327
% DeepSeek :	0.0939	0.2342	0.6720

\begin{table}[t]
    \centering
    \footnotesize
    \begin{tabular}{p{.25cm}p{6.25cm}}
        \toprule
&\textbf{\gh{}}: I \frex{pësc} \frex{mor} \frex{suvënz} \frex{per gauja dla} cunzentrazion \frex{auta} \frex{de} \frex{tuesse} tl'eghes.  \\ 
\midrule
\multirow{3}{*}{\rotatebox{90}{\textbf{\gptFourO{}\quad\;}}} &\ZS: \correct{I pësc} moro suvent per ieie dla cunsentrazion ota de tuesc tl'ega. \\
%\RS: \correct{I} pësci moran sovenz per greus alta cunzentraziun de tjess tl'egnes. \\
&\FS: \correct{I pësc mör gonot} por gauja \correct{dles} conzentrazion alta da tosser tl'ecas.\\
&\PF: \correct{I pësc} ne \correct{mör} sovenz por gauja dl cunzentrażion eveata de töies tl'egnes.\\
\midrule
\multirow{3}{*}{\rotatebox{90}{\textbf{\llama{}\quad}}} &\ZS: \correct{I} pesc se à ziplà de spess per la gauja dla concentrazion auta de tuess te \correct{eghes}. \\
%\RS: \correct{I} pesc muor de spes per via dla concentrazion auta de tosse te l'ega. \\
&\FS: \correct{I pësc mör gonot} por gauja \correct{dles} cunzentrazions altas de tosser te eglises.\\
&\PF: \correct{I} pesc muion spesso per la ragion de na cuncentrazion auta de velen tl’ega.\\
\midrule
% &\textbf{\deepseek{}} \\
\multirow{3}{*}{\rotatebox{90}{\textbf{\deepseek{}\;}}} &\ZS: \correct{I} pesc muer sovenz per via de la cunzentrazion auta de tossich tla ega.  \\
%\RS: \correct{I pësc} mor suvënz per causa dla cunzentraziun auta de tuesse tl'ega.\\
&\FS: \correct{I pësc mör gonot} por gauja dla cunzentrazion alta de tosser \correct{tles eghes}.\\
&\PF: \correct{I pësc mör gonot} por gauja dla cunzentrazion auta de tuesse \correct{tles eghes}.\\
\midrule
& \textbf{\vb{}}: I pësc mör gonot porvia dles conzentraziuns altes dla tossina tles eghes.  \\ 
\bottomrule
    \end{tabular}
    \caption{Example translation \gh{} to \vb{}.}
    \label{tab:examples-lvb-lgh}
\end{table}

\paragraph{Limited Adaptation in High-Resource Language Translation}

Our findings indicate that, in the Ladin to Italian translation direction, neither the tested prompting strategies nor higher coverage of retrieved information lead to noticeable improvements in translation quality. This observation aligns with~\citet{Robinson:etal:2023}, who noted that few-shot prompting offers limited benefits. In some cases, providing additional information even led to degraded performance, an effect also reported by \citet{alves-etal-2023-steering, deepseek:r1:2024, reynold:mcdonell:2021}. An exception is \oOne{} and \llama{}, which achieve the best results with the \FS{} method for \lvb{} to \ita{} translation.
%, possibly due to differences in overall model performance and reasoning capabilities. 
To contextualise the results, we computed the \BLEU{} score for English to Italian translation using \gptFourO{} and obtained a value of $\approx 30$ \BLEU, demonstrating strong \ZS{} performance, while still highlighting room for improvement, considering that the Ladin texts are expert translations of the original English. 
We observed notable performance differences across the models where \oOne{} often retained Ladin words. In contrast, \gptFourO{} and \deepseek{} consistently generated acceptable results without source-language interference. \gptThree{} and \llama{} sometimes adhering too closely to Ladin sentence structure. These results primarily reflect the underlying multilingual capabilities of the models.

\paragraph{NMT Models Are Superior for Translation into Low-Resource Languages} 

For Ladin, NMT models have so far only been published for the language pair \lvb{}--\ita{}~\cite{Frontull:Moser:2024}.
We have evaluated them on our test data in order to obtain a performance comparison.
For Italian to Ladin (Val Badia), the best performing NMT model (\texttt{L4}) achieved a \BLEU{} score of $16.77$, which is significantly higher than the best performing LLM. 
This confirms that while the \FS{} method allows for significant improvements, specialised NMT models remain superior for this task~\cite{Scalvini:etal:2025,Robinson:etal:2023,aycock2024llmsreallylearntranslate}. 
However, these NMT models were trained also with monolingual texts, giving them access to language-specific information that was not leveraged in our prompting experiments.
Methods that aim to incorporate such monolingual data have, for instance, been explored in~\citet{guo-etal-2024-teaching}.
In the opposite direction, the best performing NMT model \texttt{R4} achieved a \BLEU{} score of $17.04$, which is lower than the one achieved by the different LLMs with \ZS. 
This highlights the potential of LLMs for low-resource languages. For example, they can be used to produce higher-quality initial translations~\cite{Ondrejova:etal:2024}, which is a particularly relevant aspect of the widely used \emph{back-translation}~\cite{Sennrich:etal:2016}.

\section{Conclusion}

In this work, we introduced Fragment-Shot Prompting, a novel in-context learning method that improves the quality of translations into and between low-resource languages with LLMs by retrieving examples based on syntactic coverage. 
%When translating into Ladin, this approach consistently improves translation quality. 
Building on this idea, we proposed Pivoted Fragment-Shot: an extension that enables translation without the need for direct parallel data, leveraging a pivot language instead. %This method, however, is only effective with models that exhibit strong reasoning capabilities.
While prompt engineering only offers marginal improvements when translating into high-resource languages, it becomes significantly more impactful in low-resource scenarios.
Our experiments emphasise the importance of syntactic coverage in example selection. 
However, selecting examples solely based on syntactic overlap, without access to semantic information, makes it difficult to capture connections between source and target language, as fragments may have multiple meanings and uses.
As a result, effective translation in such cases requires models with strong reasoning capabilities.
At the same time, reasoning enables models to generalise beyond the examples, reducing the need for large amounts of data.
%
%By publicly releasing our code and retrieval corpora under a \texttt{CC BY-NC-SA 4.0} licence, we intend to foster further research in this area.

\paragraph{Future Work} 
Our results show that in-context learning combined with the reasoning capabilities of LLMs can improve translation quality for low-resource languages, but the mechanisms underlying this effectiveness still need to be understood in more detail~\cite{chitale-etal-2024-empirical,alves-etal-2023-steering}, highlighting the need for further research. 
Future work could also consider approaches that go beyond a single-query approach that guide the reasoning process, such as through the use of query languages like LMQL~\cite{Beurer:etal:2023}.

\section*{Ethical Statement}

We see a particular community consciousness in low-resource languages that are perceived as more trustworthy in certain contexts. For example, speakers of such languages have so far hardly been affected by phishing or similar attacks in their native language. With technological advances, these methods could be abused to exploit precisely this "greater trust" that these languages retain in the digital world.

\section*{Limitations}
Our approach augments prompts with data from an external corpus. While this method has the potential to improve machine translation for low-resource languages, we recognise several risks associated with its use. LLMs often lack robust safety mechanisms for low-resource languages, making them more prone to generating inaccurate, inappropriate or harmful content~\cite{yong2024lowresourcelanguagesjailbreakgpt4,deng2024multilingualjailbreakchallengeslarge,zeng-etal-2024-johnny}. Augmenting prompts with external data could exacerbate this problem by increasing hallucinations, potentially leading to the production of offensive or misleading translations and texts. In addition, our approach retrieves relevant ICL-examples based on syntactic similarity to fragments of the input sentence. However, this purely syntactic selection does not take into account potential biases in the training data or in the model itself. As a result, our method may inadvertently propagate or even amplify existing biases, raising concerns about fairness and ethical use. Future work should explore strategies to mitigate these risks, such as refining selection criteria beyond syntax or incorporating bias-aware filtering mechanisms.

We only conducted experiments on translation between Ladin and Italian. As Italian belongs to the same language family as Ladin and shares structural and lexical similarities, our methods may have benefited from these similarities. For more distant languages, translation may be more challenging and further evaluation is needed to assess the generalisability of our findings. 

\new{Not all prompts included an instruction on in which format the result should be returned, and even when they did, the automatic readout of the translations was not always possible. Moreover, the models occasionally generated additional content or information that went beyond the translations to be generated, e.g. to explain the generated translations or to warn the user of a lack of knowledge of the Ladin language. Since the amount of generated translations was manageable, we parsed the translations manually from the generated output. However, we note that this should be considered for efficient scaling of the experiments.}

Although the \PF-method shows promising results when translating between variants of Ladin, the size of the prompts generated by this nesting is a point of criticism. The size of the prompts increases rapidly depending on the length of the input and the translations found, as there are no assumptions about how the fragments in the source sentence correspond to those in the pivot language. The result is a search for example translations for all the fragments in the intermediate sentences. In our case, the translations in the corpus were simple, short sentences and so we have not yet reached the limits here, but this could look different with other training data. Introducing an alignment for these fragments could reduce the size of the prompts and improve efficiency by allowing less relevant examples to be omitted.

One limitation of our work is the relatively small test data set, which consists of only $175$ sentences. 
A more comprehensive evaluation using the full FLORES+ dataset would provide more robust and representative results.
To ensure compliance with the original access conditions, any release of the dataset should be complete and formally submitted to OLDI, thereby preserving its integrity as an evaluation benchmark. Given that our dataset is incomplete, we have chosen not to release it to prevent unintended uses, as it may end up in the training data of models - an outcome that would compromise FLORES+'s role as a benchmark for assessing translation quality. 

\bibliography{custom}

\end{document}